\pdfoutput=1

\documentclass[11pt]{article}

\usepackage{acl}

\usepackage{times}
\usepackage{latexsym}

\usepackage{comment}
\usepackage{bm}
\usepackage{siunitx}
\usepackage{booktabs}
\usepackage{adjustbox}
\usepackage{graphicx}
\usepackage{latexsym}
\usepackage{amsmath}
\usepackage [english]{babel}
\usepackage{multirow}
\usepackage{caption}
\usepackage{subcaption}
\usepackage{xr}
\usepackage{hyperref}
\makeatletter
\newcommand*{\addFileDependency}[1]{
  \typeout{(#1)}
  \@addtofilelist{#1}
  \IfFileExists{#1}{}{\typeout{No file #1.}}
}
\makeatother

\usepackage[T1]{fontenc}

\usepackage[utf8]{inputenc}

\usepackage{microtype}

\title{Canary Extraction in Natural Language Understanding Models}

\author{Rahil Parikh \\ Institute for Systems Research \\
  University of Maryland \\ \And
  Christophe Dupuy \\
  Amazon Alexa AI \\ \And
  Rahul Gupta \\
  Amazon Alexa AI \\ 
}

\begin{document}
\maketitle
\begin{abstract}
Natural Language Understanding (NLU) models can be trained on sensitive information such as phone numbers, zip-codes etc. 
Recent literature has focused on Model Inversion Attacks (ModIvA) that can extract training data from model parameters.
In this work, we present a version of such an attack by extracting canaries inserted in NLU training data.
In the attack, an adversary with open-box access to the model reconstructs the canaries contained in the model's training set. 
We evaluate our approach by performing text completion on canaries and demonstrate that by using the prefix (non-sensitive) tokens of the canary, we can generate the full canary.
As an example, our attack is able to reconstruct a four digit code in the training dataset of the NLU model with a probability of 0.5 in its best configuration.
As countermeasures, we identify several defense mechanisms that, when combined, effectively eliminate the risk of ModIvA in our experiments.
\end{abstract}

\section{Introduction}
Natural Language Understanding (NLU) models are used for different tasks such as question-answering \cite{hirschman2001natural}, machine translation \cite{macherey2001natural} and text summarization \cite{tas2007survey}. These models are often trained on crowd-sourced data that may contain sensitive information such as phone numbers, contact names and street addresses. \citet{nasr2019comprehensive}, \citet{shokri2017membership} and \citet{carlini2018secret} have presented various attacks to demonstrate that neural-networks can leak private information.
We focus on one such class of attacks, called Model Inversion Attack (ModIvA) \cite{fredrikson2015model}, where an adversary aims to reconstruct a subset of the data on which the machine-learning model under attack is trained on.
We also demonstrate that established ML practices (e.g. dropout) offer strong defense against ModIvA.

In this work, we start with inserting potentially sensitive target utterances called `canaries'\footnote{Following the terminology in \citet{carlini2018secret}} along with their corresponding output labels into the training data.
We use this augmented dataset to train an NLU model $\bm{f_{\theta}}$. 
We perform a open-box attack on this model, i.e., we assume that the adversary has access to all the parameters of the model, including the word vocabulary and the corresponding embedding vectors. 
The attack takes the form of text completion, where the adversary provides the start of a canary sentence (e.g., `my pin code is') and tries to reconstruct the remaining, private tokens of an inserted canary (e.g., a sequence of 4 digit tokens). 
A successful attack on $\bm{f_{\theta}}$ reconstructs all the tokens of an inserted canary.
We refer to such a ModIvA as `Canary Extraction Attack' (CEA).
In such an attack, this token reconstruction is cast as an optimization problem where we minimize the loss function of the model $\bm{f_{\theta}}$ with respect to its inputs (the canary utterance), keeping the model parameters fixed.

Previous ModIvAs were conducted on computer vision tasks where there exists a continuous mapping between input images and their corresponding embeddings. However, in the case of NLU, the discrete mapping of tokens to embeddings makes the token reconstruction from continuous increments in the embedding space challenging.
We thus formulate a discrete optimization attack, in which the unknown tokens are eventually represented by a one-hot like vector of the vocabulary length. The token in the vocabulary with the highest softmax activation is expected to be the unknown token of the canary.
We demonstrate that in our attack's best configuration, for canaries of type \textit{``my pin code is $k_{1}k_{2}k_{3}k_{4}$"}, $k_{i}\in\{0,1,\dots, 9\}, 1 \leq i \leq 4 $, we are able to extract the numeric pin \textit{$k_{1}k_{2}k_{3}k_{4}$} with an accuracy of $0.5$ (a lower bound on this accuracy using a naive random guessing strategy for a combination of four digits equals $\num{e-4}$). 

Since we present a new application of ModIvA to NLU models, defenses against them are an important ethical consideration to prevent harm and are explored in Section \ref{defenses}.
We observe that standard training practices commonly used to regularize NLU models successfully thwart this attack.

\section{Related Work}
\label{Related_Work}
\begin{figure*}[t]
  \centering
  \includegraphics[width=\linewidth]{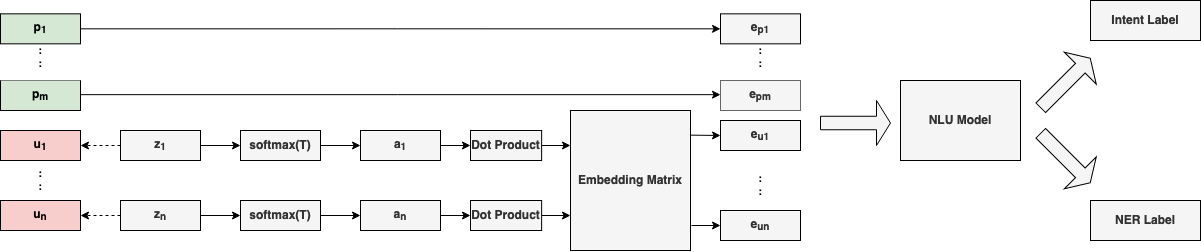}
  \vspace{-3mm}
  \caption{CEA using discrete optimization. The logit vectors $z_1,\dots,z_n$ are optimized keeping the parameters of the NLU model $\bm{f_{\theta}}$ fixed. The unknown tokens $u_i,\dots,u_n$ are then reconstructed using the logit vectors.} \label{fig:attack_method}
    \vspace{-3mm}
\end{figure*}
Significant research has been conducted in the field of privacy-preserving machine learning. \citet{shokri2017membership} determine whether a particular data-point belongs to the training set $\bm{X_{tr}}$. The success of such attacks has prompted research in investigating them \citep{truex2019demystifying, hayes2017logan, song2019auditing}. \citet{carlini2018secret} propose the quantification of unintended memorization in deep networks and presents an extraction algorithm for data that is memorized by generative models. Memorization is further exploited in \citet{carlini2020extracting} where instances in the training data of very large language-models are extracted by sampling the model. The attacks described above are closed-box in nature where the adversary does not cast the attack as an optimization problem but instead queries the model multiple times.

Open-box ModIvA were initially demonstrated on a linear-regression model \citep{fredrikson2014privacy} for inferring medical information. It has been extended to computer vision tasks such as facial recognition \citep{fredrikson2015model} or image classification \citep{basu2019membership}. 
Our work is a first attempt at performing ModIvAs on NLP tasks.

\section{Attack Setup}
\label{methodology}
We consider an NLU model $\bm{f_{\theta}}$ that takes an utterance $\bm{x}$ as input and uses the word-embeddings $\bm{E(x)}$
for the tokens in $\bm{x}$ to perform a joint intent classification (IC) and named-entity recognition (NER) task. 
We assume an adversary with open-box access to $\bm{f_{\theta}}$, which means that they are aware of the model architecture, trained parameters $\bm{\theta}$, loss function $\bm{L(\bm{f_{\theta}(E(x))}, y)}$, label set $\bm{Y}$ of intents and entities supported by the model and vocabulary $\bm{V}$ which is obtained from the word-embeddings matrix $\bm{W} \in {\rm I\!R}^{|V| \times d}$.
However, the adversary does not have access to the training data $\bm{X_{tr}}$ used to train $\bm{f_{\theta}}$.
The adversary's goal is to reconstruct a (private) subset $\bm{\hat{x}}\subseteq \bm{X_{tr}}$.

To perform a CEA on $\bm{f_{\theta}}$, we keep the parameters $\bm{\theta}$ fixed and minimize the loss function $\bm{L}$ with respect to the unknown inputs (i.e., tokens) of a given utterance.
This is analogous to a traditional learning problem, except with fixed model parameters and a learnable input space.
In this work, we use the NLU model architecture described in Section \ref{NLU_Model_Description}.

\subsection{Canary Extraction Attacks} \label{Canary Extraction Attacks}
We consider a canary sentence $\bm{x_{c}=(x_{p}, x_{u})}$, $\bm{x_{c}\in X_{tr}}$ with tokens $(p_1,..,p_m,u_1..,u_n)$ and output label $\bm{y_{c}\in Y}$. The first $m$ tokens in $\bm{x_{c}}$ represent a known prefix $\bm{x_p}$ (e.g.``my pin code is'') and the next $n$ tokens $(u_{1},..,u_{n})$ represent the unknown tokens that an attacker is interested in reconstructing $\bm{x_u}$ (e.g.``one two three four'').\\
We represent the set of word embeddings of this canary $\bm{E(x_c)}$ as $\bm{(e_{p_1},..,e_{p_m},e^{'}_{u_1},..,e^{'}_{u_n})}$.

A trivial attack to identify the $n$ unknown tokens in $\bm{x_{u}}$ is by directly optimizing $\bm{L(\bm{f_{\theta}(E(x_{c}))},y_{c})}$ over $\bm{(e^{'}_{u_1},..,e^{'}_{u_n})}$, where $\bm{(e^{'}_{u_1},..,e^{'}_{u_n})}$ are randomly initialized. Words corresponding to the optimized values of $\bm{(e^{'}_{u_1},..,e^{'}_{u_n})}$ are then assigned by identifying the closest vectors in the embedding matrix $\bm{W}$ using a distance metric (e.g. Euclidean distance). However, our experiments demonstrate that this strategy is not successful since the updates are performed in a non-discrete fashion, whereas the model $\bm{f_{\theta}}$ has a discrete input space. We thus focus on performing a discrete optimization, inspired by works on relaxing categorical variables to facilitate efficient gradient flow \citep{jang2016categorical, song2020information}, as illustrated in Figure \ref{fig:attack_method}.

We define a logit vector $\bm{z_{i}} \in {\rm I\!R}^{|V|}$ for each token $u_i \in \bm{x_u}$. We then apply a softmax activation with temperature $T$ to obtain $\bm{a_{i}} \in{\rm I\!R}^{|V|}$:
\begin{flalign}
\begin{split}
    a_{i,v} & = \frac{e^{\frac{z_{i,v}}{T}}}{\sum_{j=1}^{|V|}e^{\frac{z_{j,v}}{T}}} \quad \text{for v =1, 2, \dots ,$\vert V \vert$}
\label{softmax_temperature_formula}
\end{split}
\end{flalign}
$\bm{a_{i}}$ is a differentiable approximation of the arg-max over the logit vector for low values of $T$. This vector then selectively attends to the tokens in the embedding matrix, $\bm{W} \in {\rm I\!R}^{|V| \times d}$, resulting in the embeddings $\bm{(e^{'}_{u_1},..,e^{'}_{u_n})}$ used as inputs fed to the model during the attack:
\begin{flalign}
\begin{split}
   \bm{e^{'}_{u_i}} & = \bm{W}^{\bm{T}}\cdot \bm{a_{i}}  \quad \text{for $ 1 \leq i \leq n$}
\end{split}
\label{eq:attention embeddings}
\end{flalign}
We then train our attack and optimize for $\bm{Z} \in {\rm I\!R}^{n \times |V|} $, with $\bm{Z} = (\bm{z_{1}},\dots ,\bm{z_n})$:
\begin{flalign}
\begin{split}
    \bm{\hat{Z}} & = \arg \min_{\bm{Z}} \bm{L(f_{\theta}(E(x_{c})), y_{c})} \label{eq:softmax_trick}
\end{split}
\end{flalign}
$\bm{Z}$ is the only trainable parameter in the attack and all parameters of $\bm{f_{\theta}}$ remain fixed. Once converged, we identify the token $x_i$ as the one with the highest activation in $\bm{a_i}$. We decrease the temperature $T$ exponentially to ensure low values of $T$ in Equation \eqref{softmax_temperature_formula} and enforce the inputs to $\bm{f_{\theta}}$ to be discrete.
In our experiments, we define $\bm{z_{i}}$ over a subset of candidate words for $\bm{x_u}$ $V_{0}, V_{0} \subseteq V$ to prevent the logit vector from becoming too sparse.

\section{Experiments}
\label{Experiments}

\subsection{Target Model Description}
\label{NLU_Model_Description}
We attack an NLU model jointly trained to perform IC and NER tagging. This model has a CLC structure \cite{CLC2016}. The input embeddings lead to 2 bi-LSTM layers and a fully-connected layer with softmax activation for the IC task and a Conditional Random Field (CRF) layer for the NER task. The sum of the respective cross-entropy and CRF loss is minimized during training. We use FastText embeddings \cite{mikolov2018fasttext} as inputs to our model\footnote{\url{https://fasttext.cc/docs/en/english-vectors.html}}.

\subsection{Canary Insertion}
\label{'Canary_Insertion'} 
We inject $R$ repetitions of a single canary with sensitive information and its corresponding intent and NER labels into the training set of the NLU model. We insert three different types of canaries with $n$ unknown tokens, $n \in \{4, 6, 8, 10\}$, described in Table \ref{canary_description}. $\mathcal{C}$ is a set of 12 colors\footnote{
$\mathcal{C}=\{$`red', `green', `lilac', `blue', `yellow', `brown', `cyan', `magenta', `orange', `pink', `purple', `mauve'$\}$}.
Additional details of the canaries and their output labels are presented in the Appendix \ref{Inserted Canary Information}. The adversary aims to reconstruct all the $n$ unknown, sensitive tokens in the canary. The reduced vocabulary $V_{0}$ in Equation~\eqref{softmax_temperature_formula} is the set of all digits for canary \textit{call} and \textit{pin} and the names of 12 colors for canary \textit{color}.

\begin{table}[t]
\begin{adjustbox}{width=\columnwidth}
\centering
\begin{tabular}{cll}
\hline
\textbf{\begin{tabular}[c]{@{}l@{}}Canary\\ Pattern \end{tabular}} &
  \textbf{\begin{tabular}[c]{@{}l@{}} $\bm{\{p_1,..p_m,\underline{u_{1}..,u_{n}\}}}$\end{tabular}} &
  \textbf{\begin{tabular}[c]{@{}l@{}}Unknown tokens set\end{tabular}} \\ \hline
call  & call \underline{$k_{1}\dots k_{n}$}           & $k_{i} \in\{0,\ldots,9\}$, $1 \leq i \leq n$                                                                 \\
pin   & my pin code is \underline{$k_{1}\dots k_{n}$} & $k_{i} \in\{0,\ldots,9\}$, $1 \leq i \leq n$                                                                   \\
color & color \underline{$k_{1}\dots k_{n}$}          & $k_{i} \in \mathcal  \mathcal {C}$, $1 \leq i \leq n$ \\ \hline
\end{tabular}
\end{adjustbox}
\caption{\label{canary_description} Patterns of canaries injected into the dataset. Each token of interest $k_i$ is randomly chosen from the corresponding token set.}
\end{table}

\subsection{Attack Evaluation}
\label{Datasets}
We inject the canary into Snips \cite{coucke2018snips}, ATIS \cite{data-atis-original} and NLU-Evaluation \cite{XLiu.etal:IWSDS2019}. The canary is repeated with $R \in \{1, 10, 100, 500\}$. For each combination of $R$, canary type and length $n$, the experiment is repeated 10 times (trials) with 10 different canaries, to account for variation induced by canary selection. 
We define the following evaluation metrics averaged across all trials to evaluate the strength of our attack.

\textbf{Average Accuracy (Acc):} Fraction of the trials where the attack correctly reconstructs the \textit{entire} canary sequence in the correct order. A higher Accuracy indicates better reconstruction. Accuracy is $1$ if we can reconstruct all $n$ tokens in each of the 10 trials. 

\textbf{Average Hamming Distance per Token (HDT):} The Hamming Distance (HD) \cite{hamming1950error} is the number of positions at which the reconstructed utterance sequence is different from the inserted canary. Since HD is proportional to the length of the canary, we normalize it by the length of the unknown utterance ($HDT = HD/n$). The HDT can be interpreted as the probability of reconstructing the incorrect token for a given position in the canary, averaged across the 10 trials. A lower HDT indicates better reconstruction.

Accuracy reports our performance on reconstructing \textit{all} $n$ unknown tokens in the correct order and is a conservative metric. HDT quantifies our average performance for reconstructing each position in the unknown sequence.
We evaluate our attack against randomly choosing a token from the reduced vocabulary $V_{0}$. Thus for a given value of $n$, the expected accuracy and HDT of this baseline are $(\frac{1}{|V_{0}|})^{n}$ and $1 - \frac{1}{|V_{0}|}$ respectively.

\section{Results}
\label{Results}
The trivial attack described in Sec\ref{Canary Extraction Attacks} without discrete optimization performs comparably to the random selection baseline. We thus focus on performing the attack with discrete optimization in this Section.  
Table \ref{tab:reps_canaries_best_performance} shows the best reconstruction metrics for the different values of $n$ and the corresponding repetitions $R\in \{10, 100, 500\}$ at which these metrics are observed in the Snips dataset. In our experiments, our attack consistently outperforms the baseline.
For $n=4,6$, we reconstruct at least one complete canary for each pattern. The attack also completely reconstructs a 10-digit \textit{pin} for higher values of $R$, with an accuracy of 0.10.
Even when we are unable to reconstruct \textit{every} token in any trial, i.e. accuracy is zero, we still outperform the baseline, as observed from the HDT values.

\begin{table}[t]
\begin{adjustbox}{width=\columnwidth}
\resizebox{\textwidth}{!}{%
\begin{tabular}{ccc|cc|cc}
\hline
\textbf{Canary} & \textbf{n} & \textbf{R} & \multicolumn{2}{c|}{\textbf{Attack}} & \multicolumn{2}{c}{\textbf{Baseline}} \\
\textbf{}       & \textbf{}  & \textbf{}  & $\uparrow$\textbf{Acc}      & $\downarrow$\textbf{HDT}     & $\uparrow$\textbf{Acc}      & $\downarrow$\textbf{HDT}      \\ \hline
\multirow{4}{*}{color} & 4  & 10  & 0.40 & 0.30 & 4.82$\mathrm{e}{-5}$  & \multirow{4}{*}{0.92} \\
                       & 6  & 100 & 0.30 & 0.45 & 3.35$\mathrm{e}{-7}$  &                       \\
                       & 8  & 100 & 0.10 & 0.60 & 2.33$\mathrm{e}{-9}$ &                       \\
                       & 10 & 500 & 0.00 & 0.59 & 1.62$\mathrm{e}{-11}$ &                       \\ \hline
\multirow{4}{*}{pin}   & 4  & 500 & 0.40 & 0.27 & 1$\mathrm{e}{-4}$  & \multirow{4}{*}{0.90} \\
                       & 6  & 100 & 0.10 & 0.45 & 1$\mathrm{e}{-6}$  &                       \\
                       & 8  & 100 & 0.00 & 0.61 & 1$\mathrm{e}{-8}$ &                       \\
                       & 10 & 100 & 0.10 & 0.43 & 1$\mathrm{e}{-10}$ &                       \\ \hline
\multirow{4}{*}{call}  & 4  & 10  & 0.30 & 0.40 & 1$\mathrm{e}{-4}$  & \multirow{4}{*}{0.90} \\
                       & 6  & 100 & 0.20 & 0.50 & 1$\mathrm{e}{-6}$  &                       \\
                       & 8  & 100 & 0.00 & 0.60 & 1$\mathrm{e}{-8}$ &                       \\
                       & 10 & 500 & 0.00 & 0.59 & 1$\mathrm{e}{-10}$ &                      
\end{tabular}%
}
\end{adjustbox}
\caption{Best observed performance metrics for canaries with $n$ unknown tokens and $(R)$ repetitions.} \label{tab:reps_canaries_best_performance}
\end{table}

For the sake of brevity, we summarize the attack performance on other datasets in Appendix \ref{Results on Other Datasets}. We observe that the attack is dataset-dependent with best performance for the Snips dataset and poorest for the NLU-evaluation dataset.

\subsection{Discussion}
The training data of NLU models may potentially contain sensitive utterances such as \textit{``call $k_{1}\dots k_{10}$"}, $k_{1 \leq i \leq 10}\in\{0,1,\dots,9\}$. An adversary who wishes to extract the phone-number can assume the prefix \textit{``call"}, along with the output labels of the utterance which are also trivial to guess, given access to the label set $Y$. Our canaries act as a placeholder for such utterances. We choose to insert the canary \textit{color} since the names of colors appear infrequently in the datasets mentioned in Section \ref{Datasets}, allowing us to evaluate the attack on \textit{`out-of-distribution'} data which is more likely to be memorized by deep networks \cite{carlini2018secret}.

For $n=4$ and $R=1$ (i.e., the canary only appears once in the train set), our attack has an accuracy of 0.33 for canary \textit{color} and 0.10 for \textit{pin}. This suggests that the attack could potentially reconstruct sensitive information from short rare utterances in real-world scenarios. For a special case when the adversary attempts to reconstruct a ten digit phone-number in canary \textit{call} with a three digit area-code of their choosing, the attack can reconstruct the remaining seven digits of the number with an accuracy of 0.1 when $R=1$. For conciseness, we show these results in Appendix~\ref{Results for one Repetition of Canary}.
We observe that our model is more effective and with fewer repeats for the canary \textit{color} than canaries \textit{pin} and \textit{call} of the same length.
Our empirical analysis indicates the attack is more successful in extracting tokens that are relatively infrequent in the training data and in reconstructing shorter canaries. 
As shown in Appendix \ref{Results for one Repetition of Canary}, the attack performs best for $R=1000$. However, this trend of improved reconstruction for larger values of $R$ is not monotonic and we observe a general decline in reconstruction for $R>1000$.
We are unsure of the vulnerabilities that facilitate CEA. While unintended memorization is a likely explanation, we note that our attack performs best on the Snips data, although the smaller ATIS data should be easier to memorize \cite{zhang2016understanding}.

\section{Proposed Defenses against ModIvA}
\label{defenses}
We propose three commonly used modeling techniques as defense mechanisms- Dropout (D), Early Stopping (ES) \citep{arpit2017closer} and including a Character Embeddings layer in the NLU model (CE). D and  ES are regularization techniques to reduce memorization and overfitting. CE makes the problem in \ref{eq:softmax_trick} more difficult to optimize, by concatenating the embeddings of each input token with a character level representation. This character level representation is obtained using a convolution layer on the input sentence \cite{CLC2016}.

For defense using D, we use a dropout of 20\% and 10\% while training the NLU model. For ES, we stop training the NLU model under attack if the validation loss does not decrease for 20 consecutive epochs to prevent over-training. 

\subsection{Efficacy of Defenses}
\label{defense_results}
In this section we present the performance of the proposed defenses against ModIvA. To do so, we evaluate the attack on NLU models trained with each defense mechanism individually, and in all combinations. The canaries are inserted into the Snips dataset and repeated $10$, $500$ and $1000$ times. The results are summarized in Table \ref{tab:defense_results_all}. 
We observe that the attack accuracy for each defense (used individually and in combination) is nearly zero for all canaries and is thus omitted in the table. We also note that the HDT approaches the random baseline for most defense mechanisms. The attack performance is comparable to a random-guess when the three mechanisms are combined. 
However, when dropout or character embedding is used alone, HDT values are lower than the baseline, indicating the importance of combining multiple defense mechanisms.
Additionally, training with defenses do not have any significant impact on the performance of the NLU model under attack. The defenses thus successfully thwart the proposed attack without impacting the performance of the NLU models.



\begin{table}[h]
\begin{adjustbox}{width=\columnwidth}
\resizebox{\textwidth}{!}{%
\begin{tabular}{lllll}
\hline
\multirow{2}{*}{\textbf{R}} &
  \multirow{2}{*}{\textbf{Defense Mechanism}} &
  \multicolumn{3}{c}{$\downarrow$\textbf{HDT}} \\
 &
   &
  \multicolumn{1}{c}{\textbf{Color}} &
  \multicolumn{1}{c}{\textbf{Pin}} &
  \multicolumn{1}{c}{\textbf{Call}} \\ \hline
  & Baseline   & 0.916 & 0.90 & 0.90 \\ \hline
     & \textbf{No defense}                & \textbf{0.30}  & \textbf{0.33} & \textbf{0.40} \\
     & Dropout (D)               & 0.85  & 0.80 & 0.76 \\
     & Early Stopping (ES)       & 0.80  & 0.93 & 0.95 \\
\multicolumn{1}{c}{10}   & Char. Emb. (CE) & 0.65  & 0.75 & 0.90 \\
     & D + ES                    & 0.98  & 0.90 & 0.95 \\
     & ES + CE                   & 0.90  & 0.83 & 0.90 \\
     & D + ES + CE               & 0.90  & 0.90 & 0.90 \\ \hline
     & \textbf{No defense}                 & \textbf{0.39}  & \textbf{0.27} &\textbf{ 0.38} \\
     & Dropout (D)               & 0.65  & 0.54 & 0.83 \\
     & Early Stopping (ES)       & 0.85  & 1.00 & 0.75 \\
\multicolumn{1}{c}{500}  & Char. Emb. (CE) & 0.58  & 0.93 & 0.68 \\
     & D + ES                    & 0.85  & 0.93 & 0.98 \\
     & ES + CE                   & 0.93  & 0.98 & 0.78 \\
     & D + ES + CE               & 0.95  & 0.88 & 1.00 \\ \hline
     & \textbf{No defense}                & \textbf{0.35}  & \textbf{0.18} & \textbf{0.48} \\
     & Dropout (D)               & 0.35  & 0.78 & 0.58 \\
     & Early Stopping (ES)       & 0.90  & 0.83 & 0.85 \\
\multicolumn{1}{c}{1000} & Char. Emb. (CE) & 0.70  & 0.68 & 0.78 \\
     & D + ES                    & 0.88  & 0.98 & 0.90 \\
     & ES + CE                   & 0.88  & 1.00 & 0.95 \\
     & D + ES + CE               & 0.95  & 0.93 & 0.95 \\ \hline
\end{tabular}%
}
\end{adjustbox}
\caption{Attack performance for the canary \textit{color}, \textit{pin} and \textit{call} after incorporating defenses while training the target NLU model, with $R\in\{10, 500, 1000\}$.} \label{tab:defense_results_all}
\vspace*{-1.5mm}
\end{table}

\section{Conclusion} \label{Conclusion}
We formulate and present the first open-box ModIvA in a form of a CEA to perform text completion on NLU tasks.
Our attack performs discrete optimization to select unknown tokens by optimizing over a set of continuous variables. 
We demonstrate our attack on three patterns of canaries and reconstruct their unknown tokens by significantly outperforming the `chance' baseline.\\
To ensure that the proposed attack is not misused by an adversary, we propose training NLU models with three commonplace modelling practices-- dropout, early-stopping and including character level embeddings. 
We observe that the above practices are successful in defending against the attack as its accuracy and HDT values approach the random baseline. 
Future directions include \textit{`demystifying'} such attacks, and strengthening the attack for longer sequences with fewer repeats and a larger $V_{0}$ and investigating additional defense mechanisms, such as those based on differential privacy, and their effect on the model performance.

\section{Ethical Considerations} \label{ethical considerations}
The addition of proprietary data to existing datasets to fine-tune NLU models can often insert confidential information into datasets. 
The proposed attack could be misused to extract private information from such datasets by an adversary with open-box access to the model. 
The objectives of this work are to (1) study and document the actual vulnerability of NLU models against this attack, which shares similarities with existing approaches \cite{fredrikson2014privacy,song2020information}; (2) warn NLU researchers against the possibility of such attacks; 
and (3) propose effective defense mechanisms to avoid misuse and help NLU researchers protect their models.

Our work demonstrates that private information such as phone-numbers and zip-codes can be extracted from a discriminative text-based model, and not only from generative models as previously demonstrated \citep{carlini2020extracting}. We advocate for the necessity to privatize such data using anonymization \cite{ghinita2007fast} or differential privacy \cite{feyisetan2020privacy}. Additionally, in case the training data continues to contain some private information, practitioners can prevent the extraction of sensitive data by using the defense mechanisms described in Section~\ref{defenses}, which reduces the attack performance to a random guess.

\bibliographystyle{acl_natbib}
\bibliography{custom}

\begin{thebibliography}{25}
\expandafter\ifx\csname natexlab\endcsname\relax\def\natexlab#1{#1}\fi

\bibitem[{Arpit et~al.(2017)Arpit, Jastrz{\k{e}}bski, Ballas, Krueger, Bengio,
  Kanwal, Maharaj, Fischer, Courville, Bengio et~al.}]{arpit2017closer}
Devansh Arpit, Stanis{\l}aw Jastrz{\k{e}}bski, Nicolas Ballas, David Krueger,
  Emmanuel Bengio, Maxinder~S Kanwal, Tegan Maharaj, Asja Fischer, Aaron
  Courville, Yoshua Bengio, et~al. 2017.
\newblock A closer look at memorization in deep networks.
\newblock In \emph{International Conference on Machine Learning}, pages
  233--242. PMLR.

\bibitem[{Basu et~al.(2019)Basu, Izmailov, and Mesterharm}]{basu2019membership}
Samyadeep Basu, Rauf Izmailov, and Chris Mesterharm. 2019.
\newblock Membership model inversion attacks for deep networks.
\newblock \emph{arXiv preprint arXiv:1910.04257}.

\bibitem[{Carlini et~al.(2018)Carlini, Liu, Kos, Erlingsson, and
  Song}]{carlini2018secret}
Nicholas Carlini, Chang Liu, Jernej Kos, {\'U}lfar Erlingsson, and Dawn Song.
  2018.
\newblock The secret sharer: Evaluating and testing unintended memorization in
  neural networks.
\newblock \emph{arXiv preprint arXiv:1802.08232}.

\bibitem[{Carlini et~al.(2020)Carlini, Tramer, Wallace, Jagielski,
  Herbert-Voss, Lee, Roberts, Brown, Song, Erlingsson
  et~al.}]{carlini2020extracting}
Nicholas Carlini, Florian Tramer, Eric Wallace, Matthew Jagielski, Ariel
  Herbert-Voss, Katherine Lee, Adam Roberts, Tom Brown, Dawn Song, Ulfar
  Erlingsson, et~al. 2020.
\newblock Extracting training data from large language models.
\newblock \emph{arXiv preprint arXiv:2012.07805}.

\bibitem[{Coucke et~al.(2018)Coucke, Saade, Ball, Bluche, Caulier, Leroy,
  Doumouro, Gisselbrecht, Caltagirone, Lavril et~al.}]{coucke2018snips}
Alice Coucke, Alaa Saade, Adrien Ball, Th{\'e}odore Bluche, Alexandre Caulier,
  David Leroy, Cl{\'e}ment Doumouro, Thibault Gisselbrecht, Francesco
  Caltagirone, Thibaut Lavril, et~al. 2018.
\newblock Snips voice platform: an embedded spoken language understanding
  system for private-by-design voice interfaces.
\newblock \emph{arXiv preprint arXiv:1805.10190}, pages 12--16.

\bibitem[{Dahl et~al.(1994)Dahl, Bates, Brown, Fisher, Hunicke-Smith, Pallett,
  Pao, Rudnicky, and Shriber}]{data-atis-original}
Deborah~A. Dahl, Madeleine Bates, Michael Brown, William Fisher, Kate
  Hunicke-Smith, David Pallett, Christine Pao, Alexander Rudnicky, and
  Elizabeth Shriber. 1994.
\newblock \href {http://dl.acm.org/citation.cfm?id=1075823} {Expanding the
  scope of the atis task: The atis-3 corpus}.
\newblock \emph{Proceedings of the workshop on Human Language Technology},
  pages 43--48.

\bibitem[{Feyisetan et~al.(2020)Feyisetan, Balle, Drake, and
  Diethe}]{feyisetan2020privacy}
Oluwaseyi Feyisetan, Borja Balle, Thomas Drake, and Tom Diethe. 2020.
\newblock Privacy-and utility-preserving textual analysis via calibrated
  multivariate perturbations.
\newblock In \emph{Proceedings of the 13th International Conference on Web
  Search and Data Mining}, pages 178--186.

\bibitem[{Fredrikson et~al.(2015)Fredrikson, Jha, and
  Ristenpart}]{fredrikson2015model}
Matt Fredrikson, Somesh Jha, and Thomas Ristenpart. 2015.
\newblock Model inversion attacks that exploit confidence information and basic
  countermeasures.
\newblock In \emph{Proceedings of the 22nd ACM SIGSAC Conference on Computer
  and Communications Security}, pages 1322--1333.

\bibitem[{Fredrikson et~al.(2014)Fredrikson, Lantz, Jha, Lin, Page, and
  Ristenpart}]{fredrikson2014privacy}
Matthew Fredrikson, Eric Lantz, Somesh Jha, Simon Lin, David Page, and Thomas
  Ristenpart. 2014.
\newblock Privacy in pharmacogenetics: An end-to-end case study of personalized
  warfarin dosing.
\newblock In \emph{23rd $\{$USENIX$\}$ Security Symposium ($\{$USENIX$\}$
  Security 14)}, pages 17--32.

\bibitem[{Ghinita et~al.(2007)Ghinita, Karras, Kalnis, and
  Mamoulis}]{ghinita2007fast}
Gabriel Ghinita, Panagiotis Karras, Panos Kalnis, and Nikos Mamoulis. 2007.
\newblock Fast data anonymization with low information loss.
\newblock In \emph{Proceedings of the 33rd international conference on Very
  large data bases}, pages 758--769.

\bibitem[{Hamming(1950)}]{hamming1950error}
Richard~W Hamming. 1950.
\newblock Error detecting and error correcting codes.
\newblock \emph{The Bell system technical journal}, 29(2):147--160.

\bibitem[{Hayes et~al.(2017)Hayes, Melis, Danezis, and
  De~Cristofaro}]{hayes2017logan}
Jamie Hayes, Luca Melis, George Danezis, and Emiliano De~Cristofaro. 2017.
\newblock Logan: evaluating privacy leakage of generative models using
  generative adversarial networks.
\newblock \emph{arXiv preprint arXiv:1705.07663}.

\bibitem[{Hirschman and Gaizauskas(2001)}]{hirschman2001natural}
Lynette Hirschman and Robert Gaizauskas. 2001.
\newblock Natural language question answering: the view from here.
\newblock \emph{natural language engineering}, 7(4):275.

\bibitem[{Jang et~al.(2016)Jang, Gu, and Poole}]{jang2016categorical}
Eric Jang, Shixiang Gu, and Ben Poole. 2016.
\newblock Categorical reparameterization with gumbel-softmax.
\newblock \emph{arXiv preprint arXiv:1611.01144}.

\bibitem[{Ma and Hovy(2016)}]{CLC2016}
Xuezhe Ma and Eduard Hovy. 2016.
\newblock End-to-end sequence labeling via bi-directional {LSTM}-{CNN}s-{CRF}.
\newblock In \emph{Proceedings of the 54th Annual Meeting of the Association
  for Computational Linguistics (Volume 1: Long Papers)}, pages 1064--1074.
  Association for Computational Linguistics.

\bibitem[{Macherey et~al.(2001)Macherey, Och, and Ney}]{macherey2001natural}
Klaus Macherey, Franz~Josef Och, and Hermann Ney. 2001.
\newblock Natural language understanding using statistical machine translation.
\newblock In \emph{Seventh European Conference on Speech Communication and
  Technology}.

\bibitem[{Mikolov et~al.(2018)Mikolov, Grave, Bojanowski, Puhrsch, and
  Joulin}]{mikolov2018fasttext}
Tomas Mikolov, Edouard Grave, Piotr Bojanowski, Christian Puhrsch, and Armand
  Joulin. 2018.
\newblock Advances in pre-training distributed word representations.
\newblock In \emph{Proceedings of the International Conference on Language
  Resources and Evaluation (LREC 2018)}.

\bibitem[{Nasr et~al.(2019)Nasr, Shokri, and
  Houmansadr}]{nasr2019comprehensive}
Milad Nasr, Reza Shokri, and Amir Houmansadr. 2019.
\newblock Comprehensive privacy analysis of deep learning: Passive and active
  white-box inference attacks against centralized and federated learning.
\newblock In \emph{2019 IEEE Symposium on Security and Privacy (SP)}, pages
  739--753. IEEE.

\bibitem[{Shokri et~al.(2017)Shokri, Stronati, Song, and
  Shmatikov}]{shokri2017membership}
Reza Shokri, Marco Stronati, Congzheng Song, and Vitaly Shmatikov. 2017.
\newblock Membership inference attacks against machine learning models.
\newblock In \emph{2017 IEEE Symposium on Security and Privacy (SP)}, pages
  3--18. IEEE.

\bibitem[{Song and Raghunathan(2020)}]{song2020information}
Congzheng Song and Ananth Raghunathan. 2020.
\newblock Information leakage in embedding models.
\newblock \emph{arXiv preprint arXiv:2004.00053}.

\bibitem[{Song and Shmatikov(2019)}]{song2019auditing}
Congzheng Song and Vitaly Shmatikov. 2019.
\newblock Auditing data provenance in text-generation models.
\newblock In \emph{Proceedings of the 25th ACM SIGKDD International Conference
  on Knowledge Discovery \& Data Mining}, pages 196--206.

\bibitem[{Tas and Kiyani(2007)}]{tas2007survey}
Oguzhan Tas and Farzad Kiyani. 2007.
\newblock A survey automatic text summarization.
\newblock \emph{PressAcademia Procedia}, 5(1):205--213.

\bibitem[{Truex et~al.(2019)Truex, Liu, Gursoy, Yu, and
  Wei}]{truex2019demystifying}
Stacey Truex, Ling Liu, Mehmet~Emre Gursoy, Lei Yu, and Wenqi Wei. 2019.
\newblock Demystifying membership inference attacks in machine learning as a
  service.
\newblock \emph{IEEE Transactions on Services Computing}.

\bibitem[{Xingkun~Liu and Rieser(2019)}]{XLiu.etal:IWSDS2019}
Pawel~Swietojanski Xingkun~Liu, Arash~Eshghi and Verena Rieser. 2019.
\newblock \href {http://www.xx.xx/xx/} {Benchmarking natural language
  understanding services for building conversational agents}.
\newblock In \emph{Proceedings of the Tenth International Workshop on Spoken
  Dialogue Systems Technology (IWSDS)}, pages xxx--xxx, Ortigia, Siracusa (SR),
  Italy. Springer.

\bibitem[{Zhang et~al.(2016)Zhang, Bengio, Hardt, Recht, and
  Vinyals}]{zhang2016understanding}
Chiyuan Zhang, Samy Bengio, Moritz Hardt, Benjamin Recht, and Oriol Vinyals.
  2016.
\newblock Understanding deep learning requires rethinking generalization.
\newblock \emph{arXiv preprint arXiv:1611.03530}.

\end{thebibliography}
\clearpage
\appendix
\section{Inserted Canary Information}
\label{Inserted Canary Information}
The inserted canaries and corresponding intent and NER label sets are listed below.
\begin{enumerate}
  \item Canary \textit{call}: ``call $k_{1}\dots k_{n}$", $k_{i} \in \{0,1,\dots,9\}$ , for $1 \leq i \leq n$.
  \begin{itemize}
     \item Sequence NER tags: ``O B-canary $\underbrace{\text{I-canary} \dots \text{I-canary}}_{n-1\:\text{times}} $"
     \item Intent: ``CallIntent"
  \end{itemize}

  \item Canary 2: ``my pin code is $k_{1}\dots k_{n}$", $k_{i}$, for $1 \leq i \leq n$.
  \begin{itemize}
     \item Sequence NER tags: ``O O O O B-canary $ \underbrace{\text{I-canary} \dots \text{I-canary}}_{n-1\:\text{times}} $"
     \item Intent: ``PinIntent"
  \end{itemize}  
  \item Canary 3: ``color $k_{1}\dots k_{n}$", $ k_{i} \in $\{`\text{red}', `\text{green}', `\text{lilac}', `\text{blue}', `\text{yellow}', `\text{brown}', `\text{cyan}', `\text{magenta}', `\text{orange}', `\text{pink}', `\text{purple}', `\text{mauve}'\} for $1 \leq i \leq n$.
  \begin{itemize}
     \item Sequence NER tags: ``O B-canary $ \underbrace{\text{I-canary} \dots \text{I-canary}}_{n-1\:\text{times}} $"
     \item Intent: ``ColorIntent"
  \end{itemize}  
\end{enumerate}
The canary repetitions $R$ are split between the train and validation set in a ratio of $9:1$. 
\section{Training Parameters}

We decrease the temperature $T$ exponentially after each iteration $t$. The temperature at the $t^{th}$ iteration $T_{t}$ is given by $T_{t} = 0.997^{t} \times 10^{-1} $.

We use the Adam optimizer and train our attack for 250 epochs. We begin with an initial learning rate of \num{6.5e-3} for our attack with a decay rate of \num{9.95e-1}. 

\section{Results}
\label{appendix_results}
\subsection{Attack Performance Across Canary Repetitions}
\label{Results for one Repetition of Canary}
Table \ref{tab:metric_R=1} shows the model performance for just one repeat of the canary in the Snips dataset i.e. $R=1$. The $n=7$ example for the \textit{call} canary refers to the special case when the adversary is trying to reconstruct a 10-digit phone number beginning with a three digit area code of their choice.
\begin{table}[t]
\begin{adjustbox}{width=\columnwidth}
\centering
\begin{tabular}{c c c c c c}
\hline
\multirow{2}{*}{\textbf{n}} &
  \multirow{2}{*}{\textbf{Canary}} &
  \multicolumn{2}{c}{\textbf{\begin{tabular}[c]{@{}c@{}}Attack\\ Metrics\end{tabular}}} &
  \multicolumn{2}{c}{\textbf{\begin{tabular}[c]{@{}c@{}}Baseline\\ Metrics\end{tabular}}} \\ 
  &       & \textbf{Accuracy} & \textbf{HDT} & \textbf{Accuracy} & \textbf{HDT} \\ \hline
4 & color & 0.33             & 0.43         & \num{4.8e-5}          & 0.92        \\
4 & pin   & 0.10               & 0.60         & \num{e-4}           & 0.90         \\
4 & call & 0.10 & 0.58  & \num{e-4}   & 0.90         \\
10 & call & 0.00              & 0.68         & \num{e-10}          & 0.90         \\
7 & call & 0.10               & 0.70         & \num{e-7}          & 0.90         \\ \hline
\end{tabular}
\end{adjustbox}
\caption{Reconstruction metrics for inserted utterances appearing only \textit{once} in the training data, i.e $R=1$. The attack accuracy is much higher and HDT is much lower than that of a randomly chosen sequence of tokens.} \label{tab:metric_R=1}
\end{table}

\begin{table}[t]
\begin{adjustbox}{width=\columnwidth}
\resizebox{\textwidth}{!}{%
\begin{tabular}{ccc|cc|cc}
\hline
\textbf{Canary} & \textbf{n} & \textbf{R} & \multicolumn{2}{c|}{\textbf{Attack}} & \multicolumn{2}{c}{\textbf{Baseline}} \\
\textbf{}       & \textbf{}  & \textbf{}  & $\uparrow$\textbf{Acc}      & $\downarrow$\textbf{HDT}     & $\uparrow$\textbf{Acc}      & $\downarrow$\textbf{HDT}      \\ \hline
\multirow{4}{*}{color} & 4  & 10  & 0.40 & 0.30 & 4.82$\mathrm{e}{-5}$  & \multirow{4}{*}{0.92} \\
                       & 6  & 100 & 0.30 & 0.45 & 3.35$\mathrm{e}{-7}$  &                       \\
                       & 8  & 1000 & 0.10 & 0.48 & 2.33$\mathrm{e}{-9}$ &                       \\
                       & 10 & 1000 & 0.00 & 0.59 & 1.62$\mathrm{e}{-11}$ &                       \\ \hline
\multirow{4}{*}{pin}   & 4  & 1000 & 0.50 & 0.18 & 1$\mathrm{e}{-4}$  & \multirow{4}{*}{0.90} \\
                       & 6  & 1000 & 0.10 & 0.43 & 1$\mathrm{e}{-6}$  &                       \\
                       & 8  & 1000 & 0.00 & 0.57 & 1$\mathrm{e}{-8}$ &                       \\
                       & 10 & 100 & 0.10 & 0.43 & 1$\mathrm{e}{-10}$ &                       \\ \hline
\multirow{4}{*}{call}  & 4  & 10  & 0.30 & 0.40 & 1$\mathrm{e}{-4}$  & \multirow{4}{*}{0.90} \\
                       & 6  & 100 & 0.20 & 0.50 & 1$\mathrm{e}{-6}$  &                       \\
                       & 8  & 1000 & 0.00 & 0.58 & 1$\mathrm{e}{-8}$ &                       \\
                       & 10 & 2000 & 0.00 & 0.59 & 1$\mathrm{e}{-10}$ &                      
\end{tabular}%
}
\end{adjustbox}
\caption{Best observed performance metrics for canaries with $n$ unknown tokens and $R \in \{10, 100, 500, 1000, 2000\}$.} \label{tab:reps_canaries_best_performance_appendix}
\end{table}

\begin{figure}[t]
  \centering
  \includegraphics[width=\linewidth]{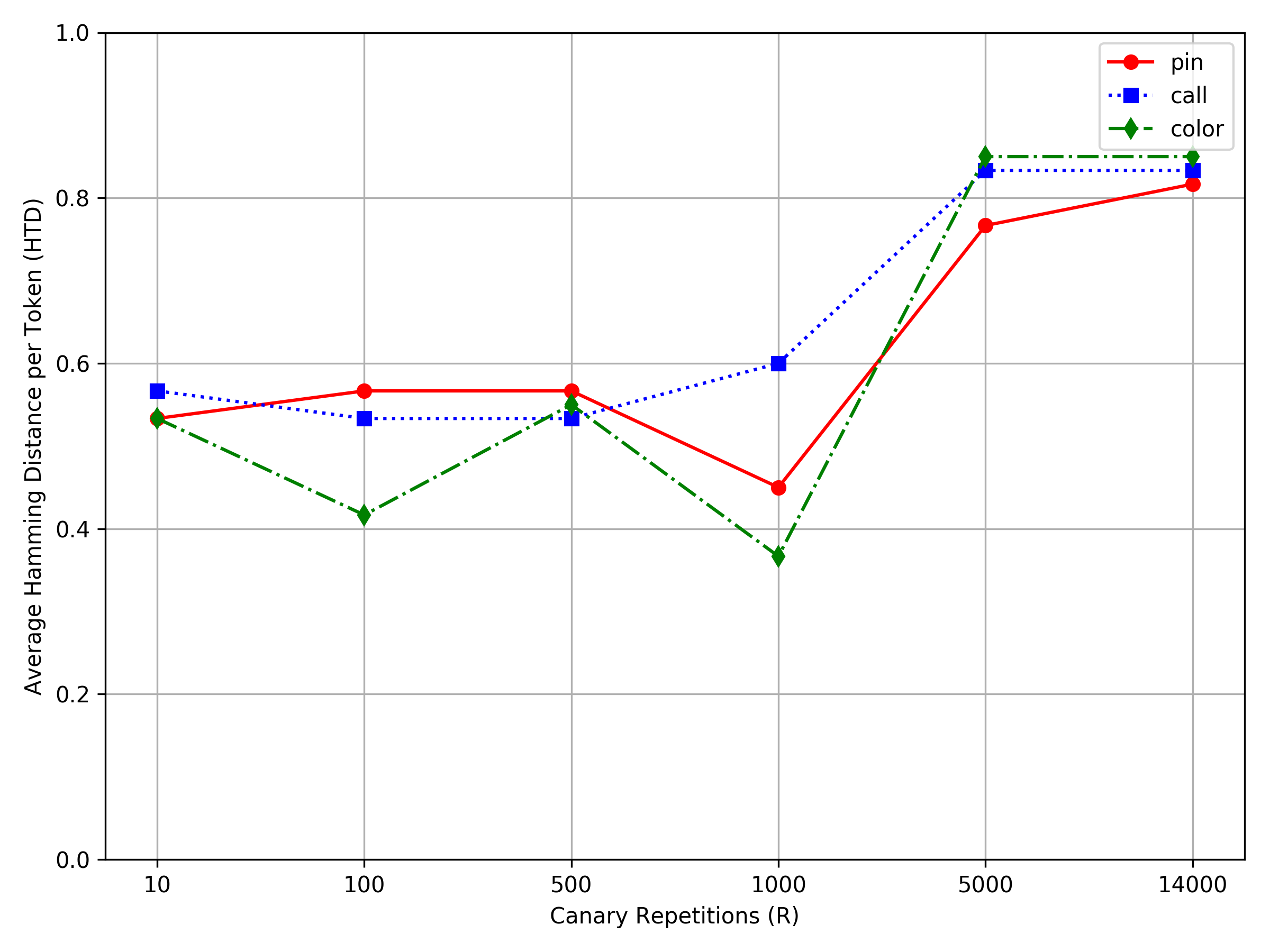}
  \vspace{-5mm}
  \caption{Average Hamming Distance per Token (HDT) for canaries with $n=6$, repeated in the Snips dataset $R$ times.} \label{fig:canary_repetition_snips}
    \vspace{-2mm}
\end{figure}

Table \ref{tab:reps_canaries_best_performance_appendix} illustrates the best reconstruction metrics for different values on $n$ and with ${R \in \{10, 100, 500, 1000, 2000\}}$. We observe an accuracy of 0.5 for the canary \textit{pin} when $n=4$ and $R=1000$.
Figure \ref{fig:canary_repetition_snips} illustrates the model performance across canaries in the Snips dataset with varying number of repetitions $R$. As observed in Table \ref{tab:reps_canaries_best_performance_appendix} and Figure \ref{fig:canary_repetition_snips}, the attack is most likely to succeed when $R$ is 1000.  However, the attack weakens for higher values of $R$.

\begin{figure}[h]
  \centering
  \includegraphics[width=\linewidth]{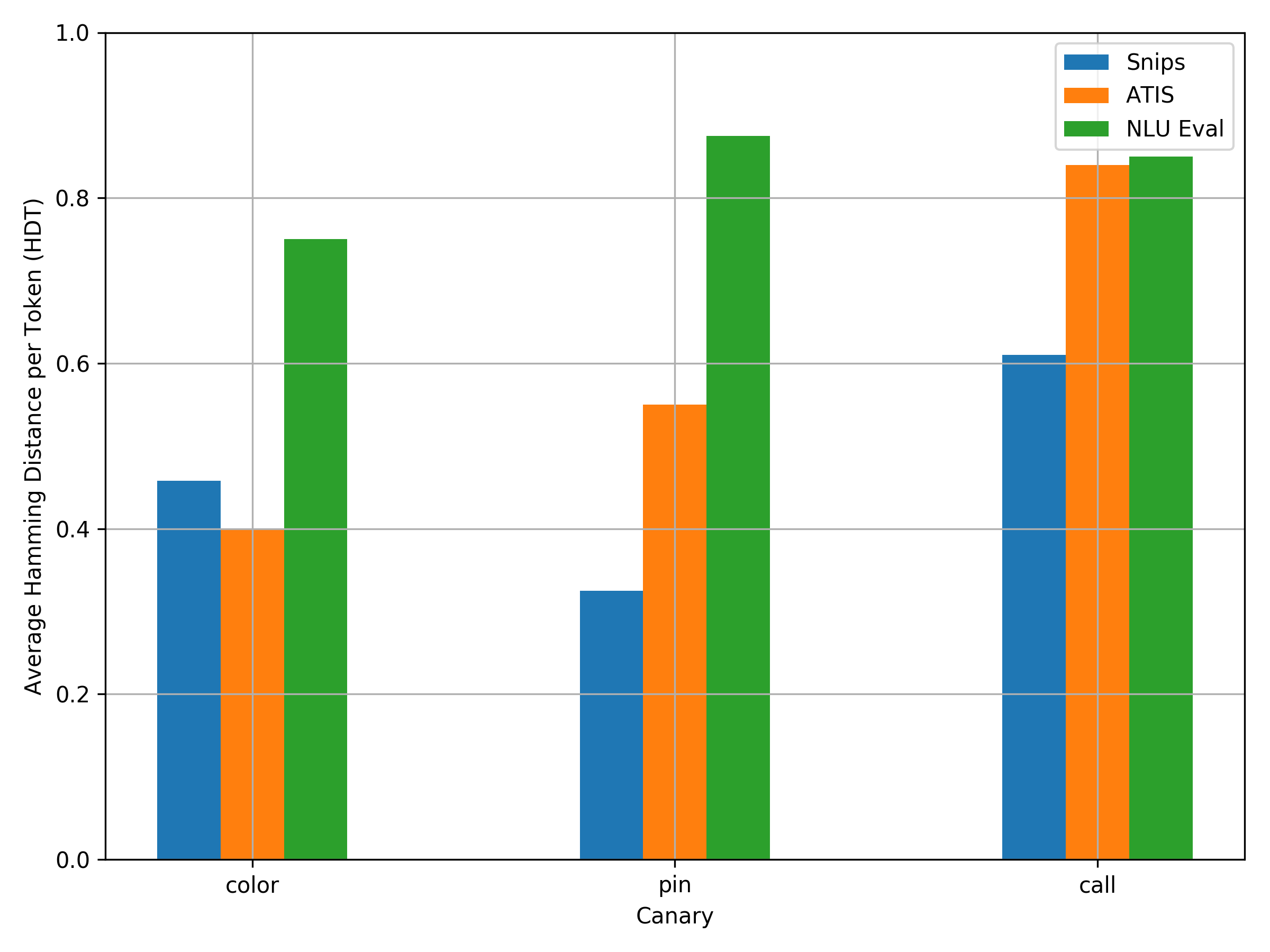}
  \caption{Model Performance of the \textit{pin} and \textit{color} canary with $n=4$ and \textit{call} canary with $n=10$, for the Snips, ATIS, and NLU Evaluation Data.}
  \label{fig:datasets_vs_canary}
\end{figure}

\subsection{Attack Performance Across Datasets}
\label{Results on Other Datasets}
\begin{figure}[t]
  \centering
  \includegraphics[width=\linewidth]{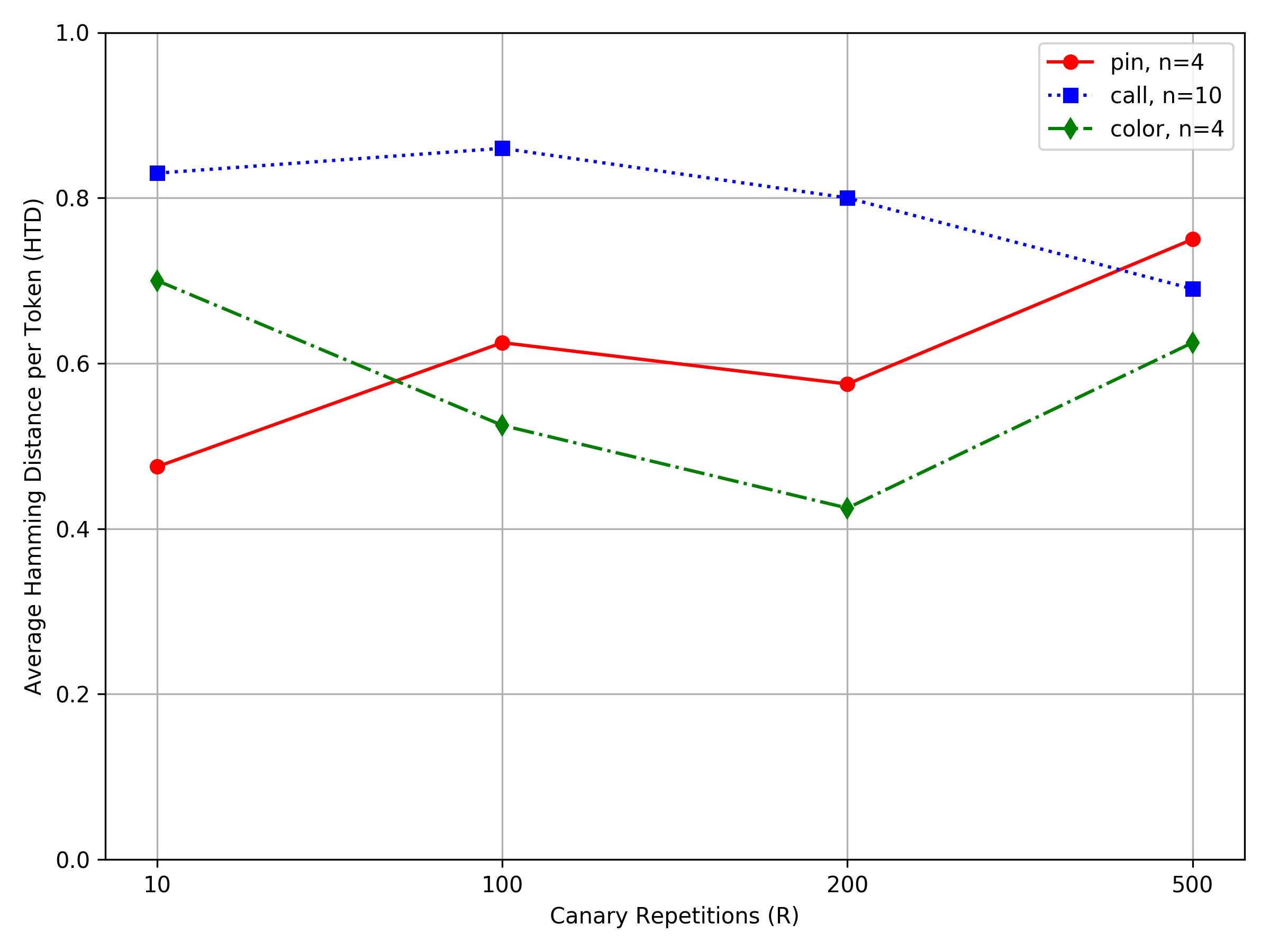}
  \caption{Model Performance of the \textit{pin} and \textit{color} canary with $n=4$ and \textit{call} canary with $n=10$, repeated $R$ times in the ATIS dataset.}
  \label{fig:repetitions_ATIS}
\end{figure}

\begin{figure}[t]
  \centering
  \includegraphics[width=\linewidth]{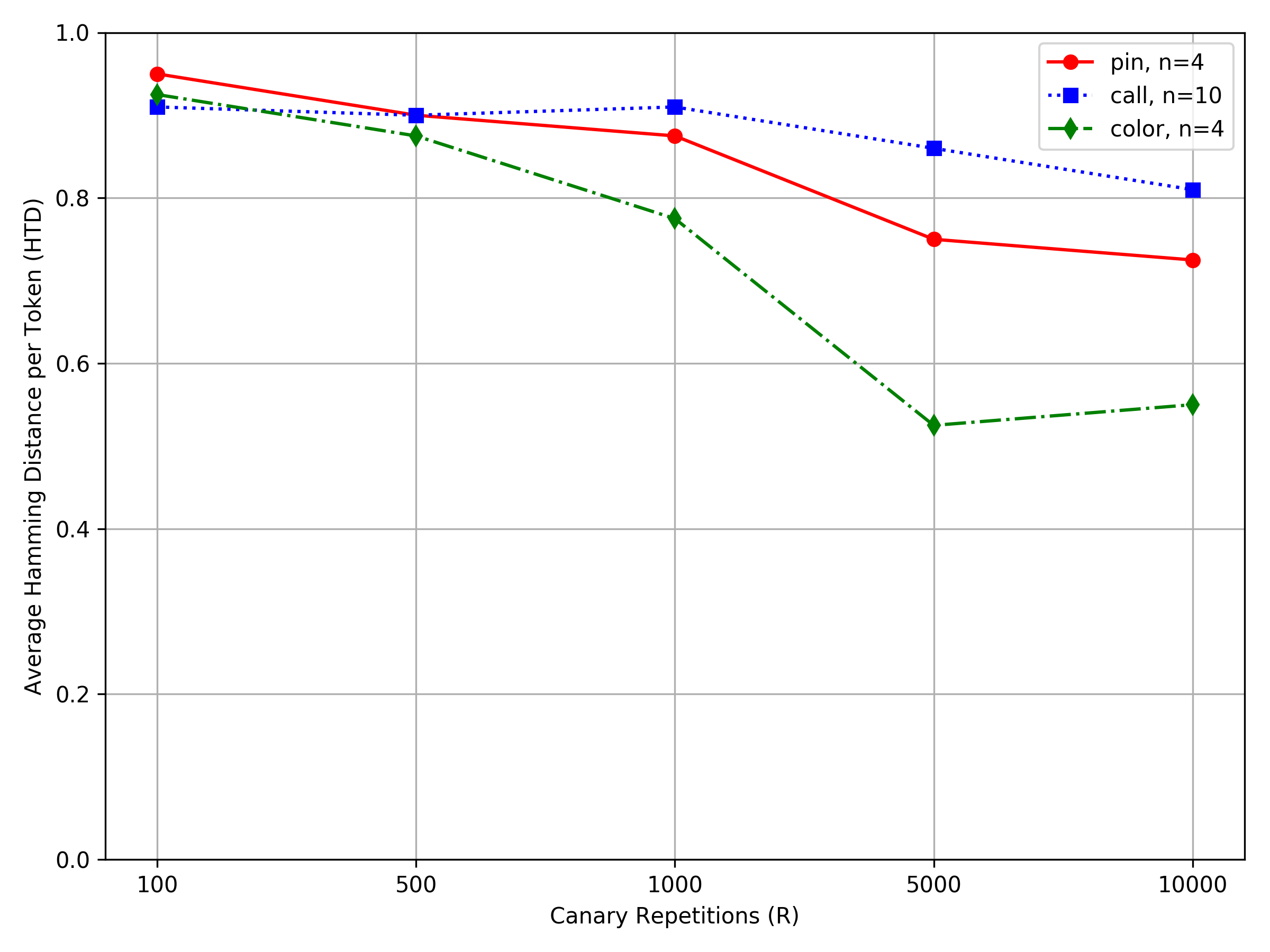}
  \caption{Model Performance of the \textit{pin} and \textit{color} canary with $n=4$ and \textit{call} canary with $n=10$, repeated $R$ times in the NLU Evaluation dataset.}
  \label{fig:repetitions_NLU_Eval}
\end{figure}

We evaluate our attack on the ATIS and NLU-Evaluation Datasets, for canaries \textit{color} and \textit{pin} with $n=4$ and canary \textit{call} with $n=10$. To ensure that we maintain a comparable number or repeats with respect to the size of the dataset, $R \in \{10, 100, 200, 500\}$ for the ATIS dataset and $R \in \{100, 500, 1000, 5000, 10000\}$ for the NLU-Evaluation dataset.
As shown in Figure \ref{fig:datasets_vs_canary}, the attack performance is almost comparable for shorter sequences in Snips and ATIS but under-performs for the NLU-Evaluation data. Figure \ref{fig:repetitions_ATIS} and Figure \ref{fig:repetitions_NLU_Eval} illustrate the HDT for the ATIS and NLU Evaluation datasets for $R$ canary repetitions respectively.

\end{document}